\def\year{2017}\relax
\documentclass[letterpaper]{article}
\usepackage{aaai17}
\usepackage{times}
\usepackage{helvet}
\usepackage{courier}
\usepackage{url}
\usepackage{latexsym}
\usepackage{graphicx}
\usepackage{multirow}
\usepackage[english]{babel}
\usepackage{color}
\usepackage{float}
\usepackage{balance}

\begin{document}
% The file aaai.sty is the style file for AAAI Press
% proceedings, working notes, and technical reports.
%
\title{Neural Machine Translation Advised by Statistical Machine Translation}
\author{Xing Wang$^\dag$ { } Zhengdong Lu$^\ddag$  { } Zhaopeng Tu$^\ddag$ { }  Hang Li$^\ddag$ { }  Deyi Xiong$^\dag$\thanks{Corresponding author}   { } Min Zhang$^\dag$ \\
  $^\dag${Soochow University, Suzhou}\\
  {\tt xingwsuda@gmail.com, \{dyxiong, minzhang\}@suda.edu.cn}\\
  $^\ddag$ {Noah's Ark Lab, Huawei Technologies, Hong Kong}\\
  {\tt \{lu.zhengdong, tu.zhaopeng, hangli.hl\}@huawei.com}
}

\maketitle
\begin{abstract}
Neural Machine Translation (NMT) is a new approach to machine translation that has made great progress in recent years.
However, recent studies show that NMT generally produces fluent but inadequate translations \cite{tu2016modeling,tu2016context,he2016improved,tu2016neural}. This is in contrast to conventional Statistical Machine Translation (SMT), which usually yields adequate but non-fluent translations.
It is natural, therefore, to leverage the advantages of both models for better translations, and in this work we propose to incorporate SMT model into NMT framework. More specifically, at each decoding step, SMT offers additional recommendations of generated words based on the decoding information from NMT (e.g., the generated partial translation and attention history). Then we employ an auxiliary classifier to score the SMT recommendations and a gating function to combine the SMT recommendations with NMT generations, both of which are jointly trained within the NMT architecture in an end-to-end manner.
Experimental results on Chinese-English translation show that the proposed approach achieves significant and consistent improvements over state-of-the-art NMT and SMT systems on multiple NIST test sets.

\end{abstract}

\section{Introduction}
\noindent Neural machine translation has been receiving considerable attention in recent years~\cite{kalchbrenner2013recurrent,cho2014learning,sutskever2014sequence,bahdanau2014neural}.
%Unlike the conventional SMT \cite{brown1993mathematics,koehn2003statistical,chiang2005hierarchical}, NMT proposes to use a single, large neural network to model the whole translation process, for which an encoder-decoder architecture is widely used.
%More specifically, the encoder encodes a variable-length source sentence into a vector and then the decoder decodes this vector to generate a variable-length target sentence.
Compared with the conventional SMT \cite{brown1993mathematics,koehn2003statistical,chiang2005hierarchical}, one great advantage of NMT is that the translation system can be completely constructed by learning from data without human involvement (cf., feature engineering in SMT). Another major advantage of NMT is that the gating \cite{hochreiter1997long} and attention \cite{bahdanau2014neural} techniques prove to be effective in modeling long-distance dependencies and complicated alignment relations in the translation process, which poses a serious challenge for SMT.

%Contrast to the conventional NMT approach which maps the whole source sentence into a fix-length vector which represents the source sentence
%\cite{kalchbrenner2013recurrent,cho2014learning,sutskever2014sequence}, Bahdanau et al. \shortcite{bahdanau2014neural} introduce the attention
%mechanism to improve the translation
%performance. The attention mechanism allows the decoder selectively focus on different parts of the source input
%to generate a context vector and has been successfully applied in many other tasks, such as image caption
%generation \cite{xu2015show}, image recognition \cite{mnih2014recurrent}, machine reading and comprehension
%\cite{hermann2015teaching},  syntactic parsing \cite{vinyals2015grammar}, etc.

Despite these benefits, recent studies show that NMT generally produces fluent yet sometimes inaccurate translations, mainly due to:
\begin{itemize}
\item[1.] {\em Coverage problem} \cite{tu2016modeling,cohn2016incorporating}: NMT lacks of a mechanism to record the source words that have been translated or need to be translated, resulting in either ``over-translation'' or ``under-translation'' \cite{tu2016modeling}.
\item[2.] {\em Imprecise translation problem} \cite{arthur2016incorporating}: NMT is prone to generate words that seem to be natural in the target sentence, but do not reflect the original meaning of the source sentence.
\item[3.] {\em UNK problem} \cite{chousing,luong2014addressing}: NMT uses a fixed modest-sized vocabulary to represent most frequent words and replaces other words with an UNK word. Experimental results  show that translation quality degrades rapidly with the number of UNK words increasing \cite{cho2014properties}.
\end{itemize}

Instead of employing different models which individually focus on the above NMT problems, in this work, we try to address the problems together in a single approach. Our approach is based on the  observation that SMT models have desirable properties that can properly deal with the above problems:
\begin{itemize}
\item[1.] SMT has a mechanism to guarantee that every source word is translated.
\item[2.] SMT treats words as discrete symbols, which ensures that a source word will be translated into a target word which has been observed at least once in the training data.
\item[3.] SMT explicitly memorizes all the translations, including translations of rare words that are taken as UNK in NMT.
\end{itemize}

It is natural to leverage the advantages of the two sorts of models for better translations. Recently, several researchers proposed to improve NMT with SMT features or outputs. For example,  He et al. (2016) integrated SMT features with the NMT model under the log-linear framework in the beam search on the development or test set. Stahlberg et al. (2016) extended the beam search decoding by expanding the search space of NMTs with translation hypotheses produced by a syntactic SMT model. In the above work, NMT was treated as a black-box and the combinations were carried out only in the testing phase.

In this work, we move a step forward along the same direction by incorporating the SMT model into the training phase of NMT. This enables NMT to effectively learn to incorporate SMT recommendations, rather than to heuristically combine two trained models.
Specifically, SMT model is firstly trained independently on a bilingual corpus using the conventional phrase-based SMT approach \cite{koehn2003statistical}.
At each decoding step, in both training and testing phases, the trained SMT model provides translation recommendations based on the decoding information from NMT, including the generated partial translation and the attention history.

We employ an auxiliary classifier to score the SMT recommendations, and use a gating function to linearly combine the two probabilities  between the NMT generations and SMT recommendations. The gating function reads the current decoding information, and thus is able to dynamically assign weights to NMT and SMT probabilities at different decoding steps.
Both the SMT classifier and gating function are jointly trained within the NMT architecture in an end-to-end manner.
In addition, to better alleviate the UNK problem in the testing phase, we select a proper SMT recommendation to replace a target UNK word by jointly considering the attention probabilities from the NMT model and the coverage information from the SMT model. Experimental results show that the proposed approach can achieve significant improvements of 2.44 BLEU point over the NMT baseline and 3.21 BLEU point over the Moses (SMT) system on five Chinese-English NIST test sets.

%\textbf{Roadmap} The rest of this paper is organized as follows, Section 2 briefly introduces the attention-based NMT and SMT as the background. Section 3 %presents our proposed model which incorporates SMT model into NMT encoder-decoder architecture. Section 4 details the training processes of the proposed model. %Section 5 presents our experiments on Chinese-English task and reports the experiment results. Finally we discuss related work and conclude the paper in Section %6 and Section 7 respectively.

\section{Background}
In this section, we give a brief introduction to NMT and phrase-based SMT, the most popular SMT model.

\subsection{Neural Machine Translation}
Given a source sentence $\mathbf{x} = x_{1}, x_{2}, ... ,  x_{T_{x}}$,  attention-based NMT \cite{bahdanau2014neural} encodes it into a sequence
of vectors, then uses the sequence of vectors to generate a target sentence $\mathbf{y} = y_{1}, y_{2}, ..., y_{T_{y}}$.

At the encoding step, attention-based NMT uses a bidirectional RNN which consists of forward RNN and backward RNN \cite{schuster1997bidirectional} to
encode the source sentence. The forward RNN reads the source sentence $\mathbf{x}$ in a forward direction, generating a sequence of forward hidden states
$\overrightarrow{h} = \left[ \overrightarrow{h_{1}}, \overrightarrow{h_{2}}, ..., \overrightarrow{h_{T_{x}}} \right]$. The backward RNN reads the source sentence $\mathbf{x}$ in a backward direction, generating a sequence of backward hidden states $\overleftarrow{h} = \left[ \overleftarrow{h_{1}}, \overleftarrow{h_{2}}, ..., \overleftarrow{h_{T_{x}}} \right]$. The pair of hidden states at each position are concatenated to form the annotation of the word at the position, yielding the annotations of the entire source sentence $\mathbf{h} = \left[ h_{1}, h_{2}, ..., h_{T_{x}} \right]$, where
\begin{equation}
%h^{\mathrm{T}}_{j} = \left[ \overrightarrow{h^{\mathrm{T}}_{j}} ; \overleftarrow{h^{\mathrm{T}}_{j}} \right]
h^{\top}_{j} = \left[ \overrightarrow{h^{\top}_{j}} ; \overleftarrow{h^{\top}_{j}} \right]
\end{equation}

At the decoding step $t$, after outputting target sequence $\mathbf{y_{<t}} = y_{1}, y_{2}, ..., y_{t-1}$, the next word $y_{t}$ is generated with probability
\begin{equation}
p(y_{t}  | \mathbf{y_{<t}}, \mathbf{x}) =  softmax(f(s_{t}, y_{t-1}, c_{t}))
\end{equation}
where $f(\cdot)$ is a non-linear activation function and $s_{t}$ is the decoder's hidden state at step $t$:
\begin{equation}
s_{t} = g(s_{t-1}, y_{t-1}, c_t)
\end{equation}
where $g(\cdot)$ is a non-linear activation function. Here we use Gated Recurrent Unit \cite{cho2014learning} as the activation function for the encoder and decoder. $c_t$ is the context vector, computed as a weighted sum of the annotations of the source sentence:
% $\mathbf{h}$
\begin{equation}
c_t = \sum_{j=1}^{T_{x}}\alpha_{t,j} h_{j}
\end{equation}
where $h_{j}$ is the annotation of source word $x_{j}$ and its weight $\alpha_{t,j}$ is computed by the attention model.
%\begin{equation}
%\alpha_{t,j} = \frac{\exp(e_{t,j})}{\sum_{k=1}^{T_{x}}\exp(e_{t,k})}
%\end{equation}
%and
%\begin{equation}
%e_{t,j} = a(s_{t-1}, h_{j})
%\end{equation}
%is an alignment model which measures how well the target word $y_{t}$ and $h_{j}$ match.

%We train the attention-based NMT model by minimizing the negative log-likelihood:
%\begin{equation}
%C(\theta) = - \frac{1}{N_{train}}\sum_{n=1}^{N_{train}} \sum_{t=1}^{T_{y}} \log p(y_{t}^{n}  | \mathbf{y_{<t}^{n}}, \mathbf{x^{n}})
%\end{equation}
%given the training data with $N_{train}$ bilingual sentences $ \{(\mathbf{x^{1}}, \mathbf{y^{1}}),(\mathbf{x^{2}}, \mathbf{y^{2}}), ..., %(\mathbf{x^{N_{train}}}, \mathbf{y^{N_{train}}}) \}$ \cite{cho2015natural}.

\subsection{Statistical Machine Translation}
Most SMT models are defined with the log-linear framework \cite{och2002discriminative}.
\begin{equation}
p(\mathbf{y} | \mathbf{x} ) = \frac{\exp(\sum_{m=1}^{M}\lambda_{m}h_{m}(\mathbf{y}, \mathbf{x}))}{\sum_{\mathbf{y}^{'}}\exp(\sum_{m=1}^{M}\lambda_{m}h_{m}(\mathbf{y}^{'}, \mathbf{x}))}
\end{equation}
where $h_{m}(\mathbf{y}, \mathbf{x})$ is a feature function and $\lambda_{m}$ is its weight. Standard SMT features include the bidirectional translation probabilities, the bidirectional lexical translation probabilities, the language model, the reordering model, the word penalty and the phrase penalty. The feature weights can be tuned by the minimum error rate training (MERT) algorithm \cite{och2003minimum}.

During translation, the SMT decoder expands partial translation (called translation hypothesis in SMT) $\mathbf{y_{<t}} = y_{1}, y_{2}, ..., y_{t-1}$  by selecting a proper target word/phrase translation for an untranslated source span from a bilingual phrase table.

\section{NMT with SMT Recommendations}

Different from attention-based NMT which predicts the next word based on vector representations, the proposed model makes the prediction also based on recommendations from an SMT model. The SMT model can be separately trained on a bilingual corpus using the conventional phrase-based SMT approach \cite{koehn2003statistical}. Given decoding information from NMT, the SMT model makes word recommendations\footnote{We have not adopted the recently proposed phraseNet which incorporates target phrases into the NMT decoder \cite{tang2016neural}. We leave the investigation to future work.} for the next word prediction with SMT features.
%These SMT recommendations can be seen as a translation guidance for the proposed model as they reflect the content of the untranslated source word/phrase.
To integrate the SMT recommendations into the proposed model, we employ a classifier to score the recommendations and a gate to combine SMT recommendations with NMT word prediction probabilities.

As shown in Figure 1, the word generation process of the proposed model has three steps:
\begin{enumerate}
\item
Inheriting from standard attention-based NMT, the NMT classifier (i.e., ``classifier$_\texttt{\scriptsize NMT}$'') estimates word prediction probabilities on the regular vocabulary $V^{nmt}$:
\begin{equation}
p_{nmt}(y_{t}  | \mathbf{y_{<t}}, \mathbf{x}) =  softmax(f(s_{t}, y_{t-1}, c_{t}))
\end{equation}
\item
The SMT classifier  (i.e., ``classifier$_\texttt{\scriptsize SMT}$'') computes probabilities of SMT recommendations, which are generated by the auxiliary SMT model.
\item
The proposed model updates word prediction probabilities by using a gating mechanism (i.e., ``gate'' in brown box).
\end{enumerate}

In the following subsections, we first elaborate on how the SMT model produces recommendations based on decoding information from the NMT model, which is the main feature of the proposed method. Then we illustrate how to integrate the SMT recommendations into NMT with an SMT classifier and a gating mechanism. Finally, we propose a novel approach to handle UNK in NMT with SMT recommendations.

%In this section we present the proposed JointDec to elaborate how to integrate the SMT model into NMT encoder-decoder architecture. Our proposed JointDec %consists of four components: 1) NMT word prediction component which inherits from attention-based NMT. 2) SMT recommendations component where JointDec
%3) Scoring Layerscores the SMT recommendations. 3) Gating mechanism which updates word prediction probabilities by using SMT recommendations and NMT word %prediction. We will detail last two components in the following subsections.

\begin{figure}[t]
\centerline{\includegraphics[width=220pt]{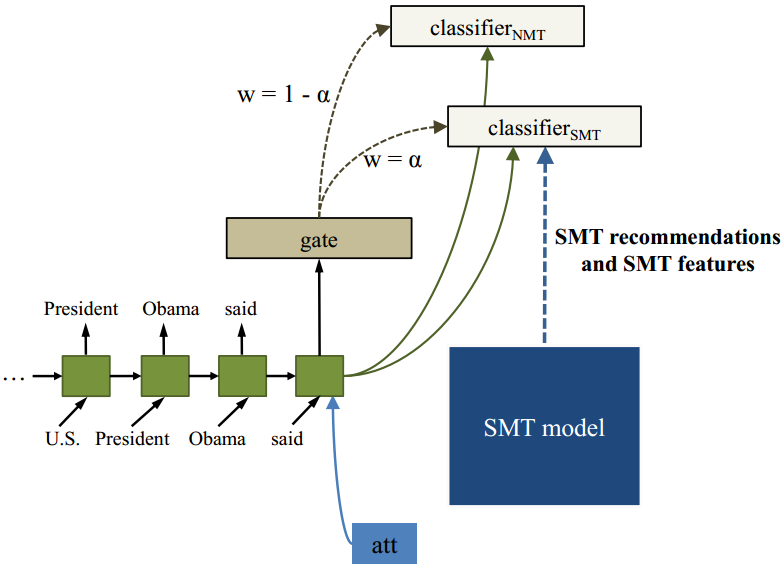}}
   % \vspace{-2em}
\caption{Architecture of decoder in NMT with SMT recommendations. The blue box with characters ``att" is the context vector outputted by the encoder of NMT with attention mechanism.}\label{figture:1}
\end{figure}

\subsection{Generating SMT Recommendations}
%SMT recommendations component undertakes two tasks: 1) SMT makes recommendations according to the previous target sequence generated by JointDec. 2) JointDec scores the SMT recommendations and generates probability estimates on them.

Given previously generated words $\mathbf{y_{<t}} = y_{1}, y_{2}, ..., y_{t-1}$ from NMT, SMT generates the next word recommendations, and computes their scores by
\begin{equation}
SMT_{score}(y_{t}  | \mathbf{y_{<t}}, \mathbf{x}) =  \sum_{m=1}^{M}\lambda_{m}h_{m}(y_{t}, x_{t})
\end{equation}
where $y_{t}$ is an SMT recommendation and $x_{t}$ is its corresponding source span. $h_{m}(y_{t}, x_{t})$ is a feature function and $\lambda_{m}$ is its weight. The SMT model can generate proper word recommendations through expanding generated words (partial translation). However there are  two problems with SMT recommendations due to the different word generation mechanisms between SMT and NMT:

\vspace{-5pt}
\paragraph{Coverage information}
SMT lacks of coverage information about the source sentence. In conventional SMT, the decoder maintains a coverage vector to indicate whether a source word/phrase is translated or not, and generates a next target word/phrase from the untranslated part of the source sentence. This coverage information is however missing since our SMT decoder has only the partial translation produced by NMT, which often leads to inappropriate recommendations due to the lack of coverage information.

To overcome this problem, we introduce an SMT coverage vector $\mathbf{cv} = \left[ cv_{1}, cv_{2}, ..., cv_{T_{x}} \right]$ for the SMT model. $cv_{i} = 1$ means that the source word $x_{i}$ has been translated and SMT will not recommend target words from translated source words. We initialize the SMT coverage vector with zeros. At each decoding step, if the output word which is generated by the proposed model is in the SMT recommendations, SMT coverage vector will be updated. Specifically, as SMT contains the positions of the recommendation words at the source sentence, we can update the SMT coverage vector by setting the corresponding vector element to 1.

\vspace{-5pt}
\paragraph{Alignment information}
The SMT reordering model lacks of alignment information between source and target sentences. Due to the word order difference between languages, the reordering model plays an import role in existing SMT approaches. Typically, the SMT model computes reordering model features based on word alignments. For example, distance-based reordering model computes the reordering cost as follows \cite{koehn2003statistical}:
\begin{equation}
d(y_{t}) = -\left| sp_{y_{t}} - sp_{y_{t-1}} - 1 \right|
\end{equation}
where $sp_{y}$ denotes the position of source word which is aligned to target word $y$.
As mentioned in the section of Background, NMT utilizes attention mechanism to align words between the target and source sides. As shown in Equation (4), instead of aligning to a specific source word, each target word is aligned to all the source words with corresponding alignment weights. Therefore the SMT reordering model can not directly work with the attention mechanism of NMT.

To address this issue, reordering cost is computed between a target word and the corresponding source words by using the alignment weights. Similar to the estimation of bi-directional word translation probabilities in \cite{he2016improved}, the reordering cost is computed as follows:
\begin{equation}
d(y_{t}) = - \sum_{j=1}^{T_{x}} \alpha_{t-1,j} \left| sp_{y_{t}} - j - 1 \right|
\end{equation}
where $\alpha_{t-1,j}$ is alignment weight produced by NMT. The lexical reordering model can work in a similar way.

\vspace{15pt}
The last question is: what kind of words does SMT recommend? Considering the second limitation of NMT described in the section of Introduction, we keep content words and filter out stop words from SMT recommendations. Now at each decoding step, with the SMT coverage vector and NMT attention mechanism, SMT can use SMT features\footnote{In training, as words out of NMT vocabulary $V^{nmt}$ may also appear in previously generated words, we should train a new language model by replacing the words with UNK and use this language model to score a target sequence.} to produce word recommendations which convey the meaning of the untranslated source word/phrase.

\subsection{Integrating SMT Recommendations into NMT}

\paragraph{SMT Classifier}
After SMT generates recommendations, a classifier scores the recommendations and generates probability estimates on them. To ensure the quality of SMT recommendations, we adopt two strategies to filter low-quality recommendations: 1) only top $N_{tm}$ translations for each source word are retained according to their translation scores, each of which is computed as weighted sum of translation probabilities\footnote{Bidirectional translation probabilities and bidirectional lexical translation probabilities.}. 2) top $N_{rec}$ recommendations with highest SMT scores are selected, each of which is computed as weighted sum of SMT features.

At the decoding step $t$, the SMT classifier takes current hidden state $s_{t}$, previous word $y_{t-1}$, context vector $c_{t}$ and SMT recommendation $y_{t}$ as input to estimate word probability of recommendation on vocabulary $V^{smt}_{t}$ . We denote the word estimated probability as:
\begin{equation}
p_{smt}(y_{t}  | \mathbf{y_{<t}}, \mathbf{x}) =  softmax(score_{smt}(y_{t}  | \mathbf{y_{<t}}, \mathbf{x}))
\end{equation}
where $score_{smt}(y_{t}  | \mathbf{y_{<t}}, \mathbf{x})$ is the scoring function for SMT recommendation defined as follows:
\begin{equation}
score_{smt}(y_{t}  | \mathbf{y_{<t}}, \mathbf{x}) = g_{smt}(f_{smt}(s_{t}, y_{t-1}, y_{t}, c_{t}))
\end{equation}
where $f_{smt}(\cdot)$ is a non-linear function and $g_{smt}(\cdot)$ is an activation function which is either an identity or a non-linear function.

As we do not want to introduce extra embedding parameters, we let SMT recommendations share the same target word embedding matrix with the NMT model. In this case, SMT word recommendation which is out of regular vocabulary $V^{nmt}$ is replaced by UNK word in the SMT classifier. Since an UNK word does not retain the meaning of the original words, SMT will not record the source coverage information if an UNK word is generated.

%\vspace{-5pt}
\paragraph{Gating Mechanism}
At last, a gate is introduced to update word prediction probabilities for the proposed model. It is computed as follows:
\begin{equation}
\alpha_{t} =  g_{gate}(f_{gate}(s_{t}, y_{t-1}, c_{t}))
\end{equation}
where $f_{gate}(\cdot)$ is a non-linear function and $g_{gate}(\cdot)$ is $sigmoid$ function.

With the help of the gate, we update word prediction probabilities on regular vocabulary $V^{nmt}$ by combining the two probabilities through linear interpolation between the NMT generations and SMT recommendations:
\begin{eqnarray}
p(y_{t}  | \mathbf{y_{<t}}, \mathbf{x}) =  (1- \alpha_{t}) p_{nmt}(y_{t}  | \mathbf{y_{<t}}, \mathbf{x}) \nonumber \\
+ \alpha_{t} p_{smt}(y_{t}  | \mathbf{y_{<t}}, \mathbf{x})
\end{eqnarray}
Note that $p_{smt}(y_{t}  | \mathbf{y_{<t}}, \mathbf{x}) = 0  \; for \; y_{t} \notin V^{smt}_{t}$.

In Equation (13), when the gate is close to 0, the proposed model will ignore the SMT recommendations and only focus on the NMT word prediction (a stop word may also be generated). This effectively allows our model to drop SMT recommendations that are irrelevant.

\subsection{Handling UNK with SMT recommendations}
To address the UNK word problem, we further directly utilize SMT recommendations to replace UNK words in testing phase.
%In testing phase, the SMT model makes recommendations as it does in training phase.
For each UNK word, we choose the recommendation with the highest SMT score as the final replacement.\footnote{There is no UNK word in the previously generated words as our model generates the target sentence from left to right. Here we can use the original language model to score the target sequence. Language model which is trained on large monolingual corpora help make accurate replacements.}
%Figure 2 shows an example of handling UNK with SMT recommendations during translation. In this case, SMT generates word recommendations and selects the word %``Rasmussen'' to replace UNK symbol.

Our method is similar to the method described in \cite{he2016improved}, both of which can make use of rich contextual information on both the source and target sides to handle UNK. The difference is that our method takes into account the reordering information and the SMT coverage vector to generate more reliable recommendations.

\section{Model Training}
We train the proposed model by minimizing the negative log-likelihood on a set of training data $\{({\bf x}^n, {\bf y}^n)\}_{n=1}^{N}$:
\begin{equation}
C(\theta) = - \frac{1}{N}\sum_{n=1}^{N} \sum_{t=1}^{T_{y}} \log p(y_{t}^{n}  | \mathbf{y_{<t}^{n}}, \mathbf{x^{n}})
\end{equation}
%given the training data with $N_{train}$ bilingual sentences $ \{(\mathbf{x^{1}}, \mathbf{y^{1}}),(\mathbf{x^{2}}, \mathbf{y^{2}}), ..., (\mathbf{x^{N_{train}}}, \mathbf{y^{N_{train}}}) \}$ \cite{cho2015natural}.

The cost function of our proposed model is the same as a conventional attention-based NMT model, except that we introduce extra parameters for the SMT classifier and the gate. We optimize our model by minimizing the cost function. In practice, we use a simple pre-training strategy to train our model. We first train a regular attention-based NMT model. Then we use this model to initialize the parameters of encoder and decoder of the proposed model and use random initialization to initialize the parameters of the SMT classifier and the gate. At last, we train all parameters of our model to minimize the cost function.

We adopt this pre-training strategy for two reasons: 1) The training of the proposed model may take a longer time with computation of SMT recommendation features and ranking of SMT recommendations, both of which are time-consuming. We treat the pre-trained model as an anchor point, the peak of the prior distribution in model space, which allows us to shorten the training time. 2) The quality of automatically learned attention weights is crucial for computing SMT reordering cost (see Equation (9)). Low quality of attention weights obtained without pre-training may produce unreliable SMT recommendations and consequently negatively affect the training of the proposed model.

\begin{table*}[t]
\centering
\begin{tabular}{l|c| c c c c c c}
\hline
SYSTEM & NIST06 & NIST02 & NIST03 & NIST04 & NIST05 & NIST08 & Avg\\
\hline
\hline
Moses                        & 32.43 & 34.08 & 34.16 & 34.74 & 31.99 & 23.69 & 31.73 \\
Groundhog                    & 30.23 & 34.79 & 32.21 & 34.02 & 30.56 & 23.37 & 30.99 \\
\hline
\hline
RNNSearch$^{*}$                & 33.53          & 35.59          & 32.55           & 36.52           & 33.05           & 24.79           & 32.50 \\
 +SMT rec                      & 34.48$^{\dag}$ & 36.40$^{\dag}$ & 34.27$^{\ddag}$ & 37.54$^{\ddag}$ & 34.25$^{\ddag}$ & 26.04$^{\ddag}$ & 33.70 \\
 +SMT rec, UNK Replace         & \textbf{34.90}$^{\ddag}$  & \textbf{38.02}$^{\ddag}$ & \textbf{36.04}$^{\ddag}$ & \textbf{38.81}$^{\ddag}$ & \textbf{35.64}$^{\ddag}$ & \textbf{26.19}$^{\ddag}$ & \textbf{34.94} \\

\hline
\end{tabular}
\caption{Experiment results on the NIST Chinese-English translation task.
%NIST06 is the development set and NIST02-05, 08 are test sets. BLEU scores in the table are case insensitive. Moses is the state-of-the-art phrase-based statistical machine translation system \cite{koehn2007moses}.
For RNNSearch, we adopt the open source system Groundhog as our baseline. The strong baseline, denoted RNNSearch$^{*}$, is our in-house NMT system. [+SMT rec, UNK Replace] is the proposed model and [+SMT rec] is the proposed model without replacing UNK words. The BLEU scores are case-insensitive. ``$\dag$'': significantly better than RNNSearch$^{*}$ ($p < 0.05$); ``$\ddag$'': significantly better than RNNSearch$^{*}$ ($p < 0.01$).}\label{table:1}
\end{table*}

\section{Experiments}
In this section, we evaluate our approach on Chinese-English machine translation.

\subsection{Setup}
The training set is a parallel corpus from LDC, containing 1.25M sentence pairs with 27.9M Chinese words and 34.5M English words\footnote{The corpus includes LDC2002E18, LDC2003E07, LDC2003E14, Hansards portion of LDC2004T07, LDC2004T08 and LDC2005T06.}. We use NIST 2006 dataset as development set, and NIST 2002, 2003, 2004, 2005 and 2008 datasets as test sets. Case-insensitive BLEU score\footnote{ftp://jaguar.ncsl.nist.gov/mt/resources/mteval-v11b.pl} is adopted as evaluation metric. We compare our proposed model with two state-of-the-art systems:
\begin{itemize}
\item[*] Moses: a state-of-the-art phrase-based SMT system with its default setting.
\item[*] RNNSearch: an attention-based NMT system with its default setting. We use the open source system GroundHog\footnote{https://github.com/lisa-groundhog/GroundHog} as the NMT baseline.
\end{itemize}

For Moses, we use the full training data (parallel corpus) to train the model and the target portion of the parallel corpus to train a 4-gram language model using the KenLM\footnote{https://kheafield.com/code/kenlm/} \cite{heafield2011kenlm}. We use the default system setting for both training and testing.

For RNNSearch, we use the parallel corpus to train the attention-based NMT model. We limit the source and target vocabularies to the most frequent 30K words in Chinese and English, covering approximately 97.7\% and 99.3\% of the data in the two languages respectively. All other words are mapped to a special token UNK. We train the model with the sentences of length up to 50 words in training data and keep the test data at the original length. The word embedding dimension of both sides is 620 and the size of hidden layer is 1000.  All the other settings are the same as in \cite{bahdanau2014neural}. We also use our implementation of RNNSearch which adopts feedback attention and dropout as NMT baseline system. Dropout is applied only on the output layer and the dropout rate is set to 0.5. We use a simple left-to-right beam search decoder with beam size 10 to find the most likely translation.

For the proposed model, we put it on the top of encoder same as in \cite{bahdanau2014neural}. As for pre-training, we train the regular attention-based NMT model using our implementation of RNNSearch and use its parameters to initialize the NMT part of our proposed model.

We use a minibatch stochastic gradient descent (SGD) algorithm together with Adadelta \cite{zeiler2012adadelta} to train the NMT models and the decay rates $\rho$ and $\epsilon$ are set as $0.95$ and $10^{-6}$. Each SGD update direction is computed using a minibatch of 80 sentences.

For SMT recommendation, we re-implement an SMT decoder which only adopts 6 features. The adopted features are bidirectional translation features, bidirectional lexical translation features, language model feature and distance-based reordering feature. Lexical reordering features are abandoned in order to speed up the SMT recommendation process. As for word-level recommendation, word penalty feature and phrase penalty feature are also abandoned as they cannot provide useful information for calculating recommendation scores. The feature weights are obtained from Moses\footnote{Compared with other weights,  the weight of distance-based reordering feature is too small, therefore we manually increase it by 10 times.}. We add English punctuations into the stop word list and remove numerals from the list. Finally the stop list contains 572 symbols. To ensure the quality of SMT recommendations, we set $N_{tm}$ to 5 and $N_{rec}$ to 25.
%At last, nearly 30 percent of words can be found in corresponding SMT recommendations in our training data.
We adopt a forward neural network with two hidden layers for the SMT classifier (Equation (11)) and gating function (Equation (12)). The numbers of units in the hidden layers are 2000 and 500 respectively.

\begin{table*}[t]
\centering
\begin{tabular}{l|c| c c c c c c}
\hline
SYSTEM & NIST06 & NIST02 & NIST03 & NIST04 & NIST05 & NIST08 & Avg\\
\hline
\hline
RNNSearch$^{*}$                     & 33.53 & 35.59 & 32.55 & 36.52 & 33.05 & 24.79 & 32.50 \\
+SMT rec                      & 34.48 & 36.40 & 34.27 & 37.54 & 34.25 & 26.04 & 33.70 \\
\hline
\hline
 $\quad +\alpha=0$                                   & 32.64 & 34.93 & 32.39 & 35.94 & 32.14 & 25.33 & 32.15 \\
 $\quad +\alpha=0.20$                                & 31.24 & 34.40 & 31.84 & 36.06 & 31.61 & 24.16 & 31.61 \\
 $\quad +$pseudo recs                                & 32.62 & 35.00 & 32.57 & 35.90 & 32.15 & 25.20 & 32.16 \\
\hline
\end{tabular}
\caption{Effect of SMT recommendations and gating mechanism.
%The model parameters are copied from RNNSearch$^{*}$$_{+SMT rec}$. The UNK symbols are unhandled in all of these models.
BLEU scores in the table are case insensitive. [+SMT rec] is the proposed model without handling UNK words. $+\alpha=0$ is [+SMT rec] with fixed gate value 0, which ignores SMT recommendations during translation. $+\alpha=0.20$ is [+SMT rec] with a fixed gate value 0.20, which ignores flexible gating mechanism. $+$pseudo recs is [+SMT rec] with pseudo SMT recommendations. }\label{table:2}
\end{table*}

\subsection{Results}

%\begin{table*}[t]
%\centering
%\begin{tabular}{l|c|c c c c c c }
%\hline
% & SYSTEM & NIST06 & NIST02 & NIST03 & NIST04 & NIST05 & NIST08 \\
%\hline
%\hline
%\multirow{2}{*}{china} & +SMT rec                                  & 150  & 86  & 43  & 158 & 100 & 59  \\
%                       & +pseudo recs                              & 470  & 271 & 254 & 437 & 324 & 377  \\
%\hline
%\hline
%\multirow{2}{*}{america} & +SMT rec                              & 4   & 5   & 11  & 28  & 13  & 14  \\
%                         & +pseudo recs                            & 231 & 123 & 165 & 322 & 219 & 271 \\
%\hline
%\end{tabular}
%\caption{Counts of ``china'' and ``america'', in [+SMT \ rec] and [+pseudo \ recs] translations. } \label{table:3}
%\end{table*}

Table 1 reports the experiment results measured in terms of BLEU score. We find that our implementation RNNSearch$^{*}$ using feedback attention and dropout outperforms Groundhog and Moses by 1.51 BLEU point and 0.77 BLEU point. RNNSearch$^{*}$$_\texttt{\scriptsize+SMT\,rec}$ using SMT recommendations but keeps target UNK words achieves a gain of up to 1.20 BLEU point over RNNSearch$^{*}$. Surprisingly, RNNSearch$^{*}$$_\texttt{\scriptsize +SMT\,rec,\,UNK\,Replace}$ which further replaces UNK word with SMT recommendation during translation, achieves further improvement over RNNSearch$^{*}$$_\texttt{\scriptsize +SMT\,rec}$ by 1.24 BLEU point. It outperforms RNNSearch$^{*}$ and Moses by 2.44 BLEU point and 3.21 BLEU point respectively.

%The key components of our proposed model JointDec are SMT recommendations and gating mechanism. So we further conducted additional experiments to validate the %effectiveness of above two components, to answer the following questions:
%\begin{itemize}
%\item[1] Are SMT recommendations indeed contributes to JointDec?
%\item[2] Does the proposed JonitDec benefit from gating mechanism?
%\item[3] Can the low-quality SMT recommendations cause a bad effect on JointDec?
%\end{itemize}

We also conduct additional experiments to validate the effectiveness of the key components of our proposed model, namely SMT recommendations and gating mechanism.  More specifically, we adopt the following three tests:  (1)  we set a fixed gate value 0 for RNNSearch$^{*}$$_\texttt{\scriptsize +SMT\,rec}$, to block SMT recommendations ($+\alpha=0$ in Table 2); (2) we set a fixed gate value 0.20 for RNNSearch$^{*}$$_\texttt{\scriptsize +SMT\,rec}$, to change the gating mechanism to a fixed mixture ($+\alpha=0.20$ in Table 2); (3) we randomly generate some high frequency target words and submit the pseudo recommendations to RNNSearch$^{*}$$_\texttt{\scriptsize +SMT\,rec}$ during translation, to deliberately confuse RNNSearch$^{*}$$_\texttt{\scriptsize +SMT\,rec}$ ($+$pseudo \ recs in Table 2). From Table 2, we can observe that:
\begin{itemize}
\item[1.] Without SMT recommendations, the proposed model suffers from degraded performance (-1.55 BLUE point). This indicates that SMT recommendations are essential for RNNSearch$^{*}$$_\texttt{\scriptsize+SMT\,rec}$.
\item[2.] Without a flexible gating mechanism, the proposed model performances on all test sets deteriorate considerably (-2.09 BLEU point). This shows that gating mechanism plays an important role in RNNSearch$^{*}$$_\texttt{\scriptsize+SMT\,rec}$.
\item[3.] The experiment results in Table 2 empirically show that the proposed model makes wrong translations and has a significant decrease in performance (- 1.54 BLEU point), which demonstrates that the quality of SMT recommendations is also very important for RNNSearch$^{*}$$_\texttt{\scriptsize+SMT\,rec}$.
\end{itemize}

\section{Related Work}
In this section we  briefly review previous studies that are related to our work. Here we divide previous work into three categories:
% 1) Combination of SMT and NMT, 2) Coverage problem in NMT, and 3) UNK word problem in NMT.
\\
\\
\noindent \textbf{Combination of SMT and NMT:} Stahlberg et al. (2016) extended the beam search decoding by expanding the search space of NMT with translation hypotheses produced by a syntactic SMT model. He et al. (2016) enhanced NMT system with effective SMT features. They integrated three SMT features, namely translation model, word reward feature and language model, with the NMT model under the log-linear framework in the beam search. Arthur et al. (2016) proposed to incorporate discrete translation lexicons into NMT model. They calculated lexical predictive probability and integrated the probability with the NMT model probability to predict the next word. Wuebker et al. (2016) applied phrase-based and neural models to complete partial translations in interactive machine translation and find the models can improve next-word suggestion. The significant difference between our work and these studies is that NMT is treated
as a black-box in the previous work, while in our work the NMT and SMT models are tightly integrated with the execution of the former being advised by the latter in the training and testing phase.
\\
\\
\noindent \textbf{Coverage problem in NMT:} Tu et al. (2016b) proposed a coverage mechanism for NMT to alleviate the ``over-translation'' and ``under-translation'' problems. They introduced a coverage vector for the attention model, to make the NMT model consider more about untranslated source words in the target word generation. Cohn et al. (2016) enhanced the attention model with structural biases from word based alignment models, including positional bias, Markov conditioning, fertility and agreement over translation directions. Feng et al. (2016) proposed to add implicit distortion and fertility models to attention model. These studies tackle the coverage problem by enhancing the encoder of NMT. As we incorporate SMT model into the decoder part of NMT, our work is complementary to the above studies.
\\
\\
\noindent \textbf{UNK word problem in NMT:} Luong et al. (2015) proposed several approaches to track the source word of an UNK word in the target sentence. They first attached aligned information for each target UNK word in training data and trained a model on the data, and then they used the model to generate translation with UNK alignment information. Jean et al. (2015) proposed an efficient approximation based on importance sampling on the softmax function which allows to train NTM with very large vocabulary. On the other hand, instead of modeling word unit, some work focused on smaller unit modeling \cite{chitnis2015variable,sennrich2015neural}, especially on character modeling \cite{ling2015character,costa2016character,luong2016achieving,chung2016character}.
Our work is also motivated by Copynet \cite{gu2016incorporating}, which incorporates copying mechanism in sequence-to-sequence learning.

\section{Conclusion}
In this paper, we have presented a novel approach that incorporates SMT model into NMT with attention mechanism. Our proposed model remains the power of end-to-end NMT while alleviating its limitations by utilizing recommendation from SMT for better generation in NMT. Different from prior work which usually used a separately trained NMT model as an additional feature, our proposed model containing NMT and SMT is trained in an end-to-end manner. Experiment results on Chinese-English translation have demonstrated the efficacy of the proposed model.

\section{Acknowledgments}
This work was supported by the National Natural Science Foundation of China (Grants No.61525205, 61432013 and 61622209). We would like to thank three anonymous reviewers for their insightful comments.

% include your own bib file like this:
%\bibliographystyle{acl}
%\bibliography{acl2014}

\balance
\nocite{feng2016implicit}
\nocite{stahlberg2016syntactically}
\nocite{meng2015deep}
\nocite{joe2016models}
\nocite{tu2016context}
\nocite{tu2016neural}
\bibliographystyle{aaai}
\bibliography{formatting-instructions-latex}
\end{document}